\documentclass{article}

\usepackage[english]{babel}

\usepackage{subcaption}

\usepackage[letterpaper,top=2cm,bottom=2cm,left=3cm,right=3cm,marginparwidth=1.75cm]{geometry}

\usepackage{float}
\usepackage{amsmath}
\usepackage{graphicx}
\usepackage[colorlinks=true, allcolors=blue]{hyperref}
\usepackage[capitalise,noabbrev]{cleveref} 

\usepackage{tabularx} %
\usepackage{lipsum} %
\usepackage{booktabs} %

\usepackage{tabularx} %
\usepackage{booktabs} %
\usepackage{makecell} %

\title{Generating Print-Ready Personalized AI Art Products from Minimal User Inputs}

\author{Noah Pursell \\ University of Oklahoma \\ \texttt{noah.a.pursell-1@ou.edu}
   \and Anindya Maiti \\ University of Oklahoma \\ \texttt{am@ou.edu}}
\date{\today}

\begin{document}
\maketitle

\begin{abstract}

We present a novel framework to advance generative artificial intelligence (AI) applications in the realm of printed art products, specifically addressing large-format products that require high-resolution artworks. The framework consists of a pipeline that addresses two major challenges in the domain: the high complexity of generating effective prompts, and the low native resolution of images produced by diffusion models. By integrating AI-enhanced prompt generations with AI-powered upscaling techniques, our framework can efficiently produce high-quality, diverse artistic images suitable for many new commercial use cases. Our work represents a significant step towards democratizing high-quality AI art, opening new avenues for consumers, artists, designers, and businesses.
\end{abstract}

\section{Introduction}

\begin{figure}[b]
\centering
\includegraphics[width=0.99\textwidth]{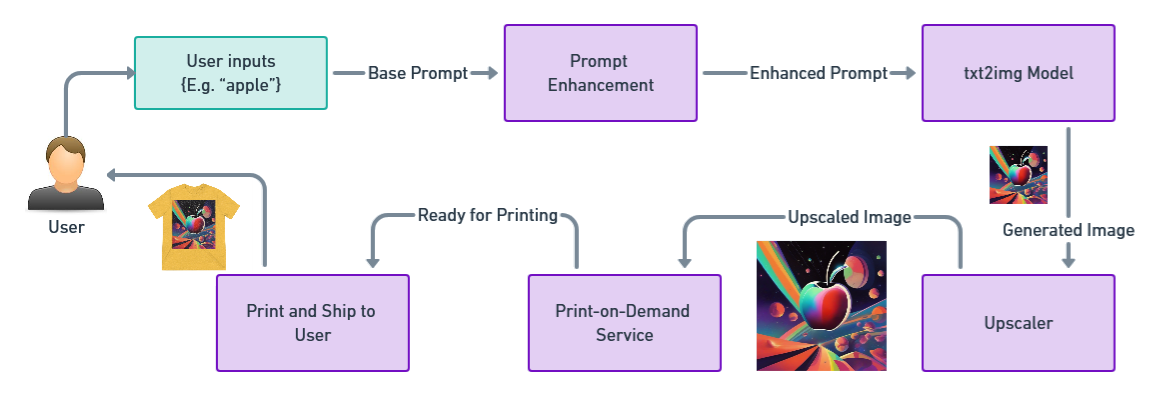}
\caption{Overview of our framework for creating printed AI art products, with minimal user inputs.}
\label{fig:overview-pipeline}
\end{figure}

In the evolving landscape of digital art, the advent of Artificial Intelligence (AI) has ushered in a groundbreaking era of creativity and innovation. Among the myriad of AI applications, AI-generated imagery \cite{ramesh2021zero,rombach2022high,sdxlhug,midjourney} stands at the forefront, transforming the conception and production of art. This revolution not only redefines traditional artistic practices but also introduces unprecedented challenges and opportunities for artists, designers, and businesses alike. This paper delves into the development of a framework for printed AI art products (\cref{fig:overview-pipeline}), focusing on overcoming two primary hurdles: the intricacy of generating effective prompts and the limitations of native image resolution in diffusion models. By synergizing advanced prompt generation methods with AI-powered upscaling techniques, this research presents a scalable solution for producing high-quality, high-resolution artistic images with minimal user inputs. Using the state-of-the-art Stable Diffusion XL model \cite{sdxlhug}, we explore the potential of AI-generated art for commercial use, demonstrating how this innovative approach can democratize art creation, enhance aesthetic appeal, and set new benchmarks for the integration of AI in artistic innovation.

\section{Background and Motivation}

In the intersection of high-resolution digital art and print commerce, platforms like Society6\footnote{https://society6.com} and Redbubble\footnote{https://www.redbubble.com} exemplify the necessity for large, detailed images to ensure product quality. For instance, art prints demand resolutions up to 6500$\times$6500 pixels, ensuring that when the artwork is printed on large canvases, the clarity and integrity of the image are maintained. Similarly, home decor items like duvet covers require 7632$\times$6480 pixels to guarantee that the designs retain their vibrancy and detail when printed. This emphasis on high resolution is pivotal in the context of AI-generated art, where the digital-to-physical transition must uphold the artist’s vision with fidelity.

In contrast, state-of-the-art image generation tools often face limitations in output resolution, typically peaking at 1024 pixels on a side, which may suffice for common digital displays but falls short for high-quality print demands on larger surfaces. This discrepancy poses a significant challenge for artists and designers aiming to bridge the gap between digital detail and physical product excellence. Hence, the advent of advanced upscaling methods becomes crucial, enabling the enhancement of native resolution to meet the stringent specifications of print-on-demand services and ensuring that the vibrancy and nuance of digital art are preserved when translated to tangible merchandise.

\subsection{The Challenge of Prompt Engineering}
Prompt engineering, the process of crafting textual inputs to guide AI models in generating specific visual outputs, poses a significant challenge in the realm of AI-generated art. This task is intricate and time-consuming, requiring a blend of creativity, technical insight, and iterative experimentation \cite{witteveen2022investigating}. For consumers looking to leverage AI for art creation, the steep learning curve and the effort involved in mastering prompt engineering can detract from the overall experience. Many users find themselves overwhelmed by the complexity of formulating effective prompts, a hurdle that can stifle creativity and accessibility. Consequently, there is a pressing need for a prompt enhancement system that simplifies this process. By automating and refining the prompt generation process, such a system would lower the barrier to entry for consumers, enabling them to easily create high-quality prompts and, by extension, high-quality art. This innovation has the potential to transform the user experience, making AI art creation more accessible and enjoyable for a wider audience.

\subsection{Limitations of Native Resolution in AI Models}
The quest for high-resolution images in AI-generated art is not merely a pursuit of visual quality but a critical requirement for commercial viability, especially in the context of art prints where resolutions around 4k are considered standard. Despite the advancements in traditional pixel-based diffusion models and Latent Diffusion Models (LDMs) \cite{rombach2022high}, including Stable Diffusion, which are capable of generating intricate and diverse imagery, these models are inherently limited by their native resolution, typically limited to 1024 pixels. This limitation highlights a significant gap between the current capabilities of AI models and the resolution demands of the commercial art market. To address this, there is a pressing need to develop and apply advanced upscaling techniques that can extend the resolution of AI-generated images far beyond the native output of these models, ensuring that the artworks not only achieve but surpass the quality required for high-end art prints and commercial applications.

\section{Proposed Solutions}

\subsection{Enhanced Prompt Generation}

First, we present methods to refine the process of prompt engineering for applying generative AI in art creation. Three distinct methodologies aimed at enhancing the prompts fed into language models to produce higher-quality outputs are presented in this section. These methods take in a \textit{base prompt} from the user, which can be a very simple combination of words, and enhance it into an \textit{enhanced prompt} which is more likely to generate visually pleasing art.

Method 1, \textbf{Language Model (LLM) Based Generation}, takes the simple approach of supplying a LLM with a base prompt, and instructions to enhance it. Various instructions can produce various results.
Method 2, \textbf{LLM with RAG-Based Multishot}, integrates Retrieval-Augmented Generation (RAG) \cite{lewis2020retrieval} with language models to further enhance prompts. Similarly to the previous method, an LLM is supplied with a base prompt and instructions to enhance it. Additionally, RAG is used to identify relevant example prompts from a pre-populated database of high-quality prompts which are given to the LLM for reference.
Method 3, \textbf{RAG-Based Templating}, utilizes RAG to harness the power of templates in prompt generation. This method uses vector similarity to identify templates, from a pre-populated database of prompt templates, that are contextually similar to the base prompt. By embedding the base prompt within a positive template the process culminates in a final enhanced prompt that is likely to generate high-quality content.

These three methods can eliminate the intricacy involved in crafting effective prompts (for an end user) and automate the process of high-quality art generation, making the creation of AI-generated art more accessible and user-friendly.

\subsubsection{Method 1: Language Model (LLM) Based Generation}

\begin{figure}[h]
\centering
\includegraphics[width=0.8\textwidth]{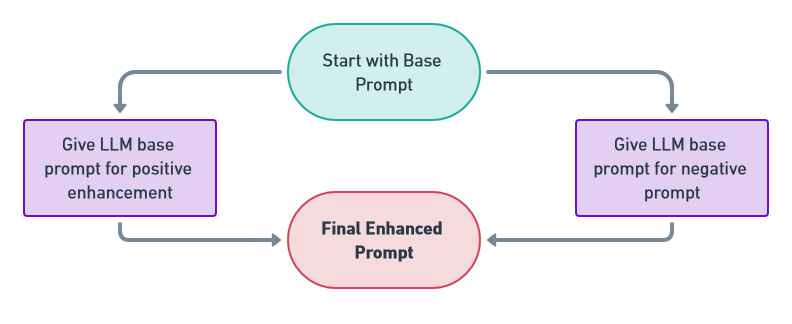}
\caption{Overview of Language Model (LLM) Based Generation.}
\label{fig:PromptEnhancementDiagrams-llm}
\end{figure}

For generating an enhanced prompt using a Language Model (LLM), starting with a base prompt and processing it through both positive and negative enhancement pathways (\cref{fig:PromptEnhancementDiagrams-llm}):

\begin{itemize}
    \item \textbf{Start with base prompt:} The process begins with a minimal one or two words typed by the user or randomly generated as textual input forming the base prompt.

    \item \textbf{Give LLM base prompt for positive enhancement:} On one pathway, the base prompt is provided to the LLM, which is instructed to generate a \textit{positive prompt} - a prompt that explains what the image generation should display. The LLM is instructed to use proven prompt engineering techniques, such as providing lighting keywords \cite{witteveen2022investigating}. 

    \item \textbf{Give LLM base prompt for negative prompt:} In the parallel process, the base prompt is also provided to the LLM for the generation of a \textit{negative prompt}. This negative prompt specifies what should be excluded from the image, such as watermarks.

    \item \textbf{Final Enhanced Prompt:} The outputs from both the positive and negative enhancement processes are used as the prompt for the image generation model.

\end{itemize}

\subsubsection{Method 2: LLM with RAG-Based Multishot}

\begin{figure}[h]
\centering
\includegraphics[width=0.8\textwidth]{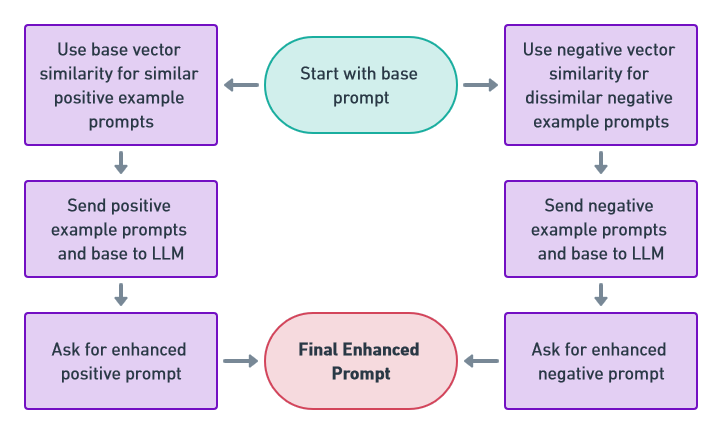}
\caption{Overview of LLM with RAG-Based Multishot.}
\label{fig:my_label}
\end{figure}

For enhancing a base prompt using LLM with RAG-Based Multishot, the following is a breakdown of the process:

\begin{itemize}
    \item \textbf{Start with base prompt:} The process begins with minimal textual input as the base prompt.

    \item \textbf{Use base vector similarity for similar positive example prompts:} On one path, the system uses vector similarity measures to identify similar positive prompts that are semantically similar to the base prompt. 

    \item \textbf{Send positive example prompts and base to LLM:} These similar positive prompts, along with the base prompt, are then sent to the LLM. 

    \item \textbf{Ask for enhanced positive prompt:} The LLM is given instructions to enhance the base prompt, using the example positive prompts as reference examples.

    \item \textbf{Use negative vector similarity for dissimilar negative example prompts:} In a parallel path, the system uses vector similarity to identify prompts that are semantically dissimilar to the base prompts. These are prompts that are different from the base prompt and describe images that are significantly different from the base prompt. These negative prompts can come from the same database as the positive prompt examples or can be separated. 

    \item \textbf{Send negative example prompts and base to LLM:} A list of dissimilar prompt examples, labeled as negative prompts, and the base prompt are sent to the LLM.

    \item \textbf{Ask for enhanced negative prompt:} The LLM is given instructions to generate a new negative prompt for the base prompt, using the example negative prompts as reference examples. 

    \item \textbf{Final Enhanced Prompt:} The outputs from both the positive and negative enhancement processes are used as the prompt for the image generation model.

\end{itemize}

\subsubsection{Method 3: RAG-Based Templating}

\begin{figure}[h]
\centering
\includegraphics[width=0.8\textwidth]{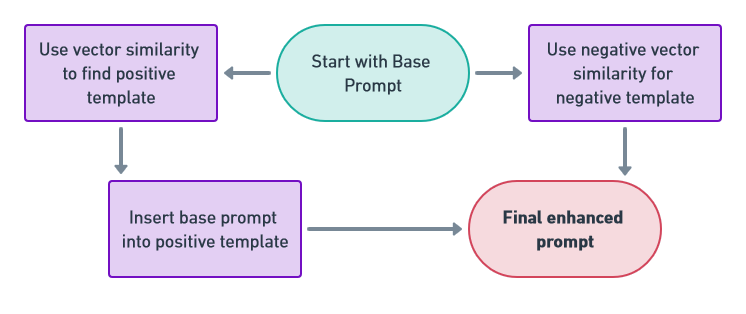}
\caption{Caption of the figure.}
\label{fig:my_label}
\end{figure}

Enhancing a base prompt using RAG-Based Templating again involves the use of Retrieval-Augmented Generation for template-based prompt enhancement. Here’s what each step implies:

\begin{itemize}
    \item \textbf{Start with base prompt:} The process begins with minimal textual input as the base prompt.

    \item \textbf{Use vector similarity to find positive template:} On one side of the process, vector similarity is used to retrieve a list of similar positive templates. One of these is randomly selected via a weighted random selection. 

    \item \textbf{Insert base prompt into positive template:} The base prompt is then inserted into this positive template.

    \item \textbf{Use negative vector similarity for negative template:} In parallel, vector similarity is also used to retrieve a list of dissimilar negative prompts. One of these is randomly selected via a weighted random selection.  

    \item \textbf{Final enhanced prompt:} The outputs from both the positive and negative enhancement processes are used as the prompt for the image generation model.

\end{itemize}

\subsection{Advanced Upscaling Algorithms}

Upscaling images is essential for improving the resolution and quality of digital images, particularly when enlarging them for various applications, such as art prints, without losing clarity. We analyzed nine different conventional and AI-based upscalers for our use case while considering various costs, performance metrics, and image quality factors.

\begin{itemize}
  \item \textbf{ESRGAN-4X}: Enhanced Super-Resolution Generative Adversarial Networks \cite{wang2018esrgan}, a 4x upscaler.
  \item \textbf{LDSR}: Latent Diffusion Super Resolution (LDSR) leverages on Stable Diffusion latent upscaler model.
  \item \textbf{R-ESRGAN-4X+}: A variation of ESRGAN with improved features and performance.
  \item \textbf{ScuNET-GAN}: Combines ScuNET, a network architecture for super-resolution, with GAN (Generative Adversarial Network).
  \item \textbf{ScuNET-PSNR}: Similar to ScuNET-GAN but focuses on maximizing PSNR (Peak Signal-to-Noise Ratio), a measure of image quality.
  \item \textbf{SwinIR-4X}: Refers to a super-resolution method that utilizes Swin Transformer architectures at a 4x upscaling factor.
  \item \textbf{Lanczos}: An algorithm for image resizing that uses a sinc-based kernel.
  \item \textbf{NEAREST}: Nearest Neighbor interpolation is the simplest upscaling method. Each pixel in the upscaled image is assigned the value of its nearest neighbor in the original image. It's fast but tends to produce blocky and pixelated results.
  \item \textbf{HAT-L}: The HAT-L model, referring to a Hybrid Attention Transformer \cite{chen2023activating} with a ``Large'' configuration, is an innovative approach that merges the strengths of channel attention mechanisms and window-based self-attention schemes. 
\end{itemize}

\section{Experiments}
\subsection{Image Generation Using Enhanced Prompts}

\begin{itemize}
    \item \textbf{Initial Prompt Selection:} The experiment commenced with the selection of 5 base prompts. These were chosen to represent a diverse set of themes while also being simplistic to assess the effectiveness of the enhancement techniques across various contexts
    \item \textbf{Prompt Enhancement:} Each of the previously mentioned prompt enhancement strategies was applied to all five base prompts, resulting in three sets of five enhanced prompts. 
    \item \textbf{Image Generation:} Using Stable Diffusion XL, images were generated with each of the prompts at 1024x1024 resolution. 
\end{itemize}

\subsection{Image Upscaling}
\begin{itemize}
    \item \textbf{Selection of Images for Uscaling:} From the pool of images generated in 4.1, one image per base prompt was selected, resulting in a total of five images for further analysis. The selection criteria focused on image complexity and diversity to assess the effectiveness of different upscalers on a variety of image types. 
    \item \textbf{Application of Upscaling Techniques:} Each of the previously mentioned upscalers was applied to the selected images to achieve a 4x upscaling to 4096x4096. During the upscaling process, GPU load and GPU VRAM usage were recorded using NVIDIA's PyNVML package. 
\end{itemize}

\section{Overall Impressions}

\subsection{Prompt Enhancement Analysis}

\begin{figure}[H]
    \centering
    \includegraphics[width=0.9\textwidth]{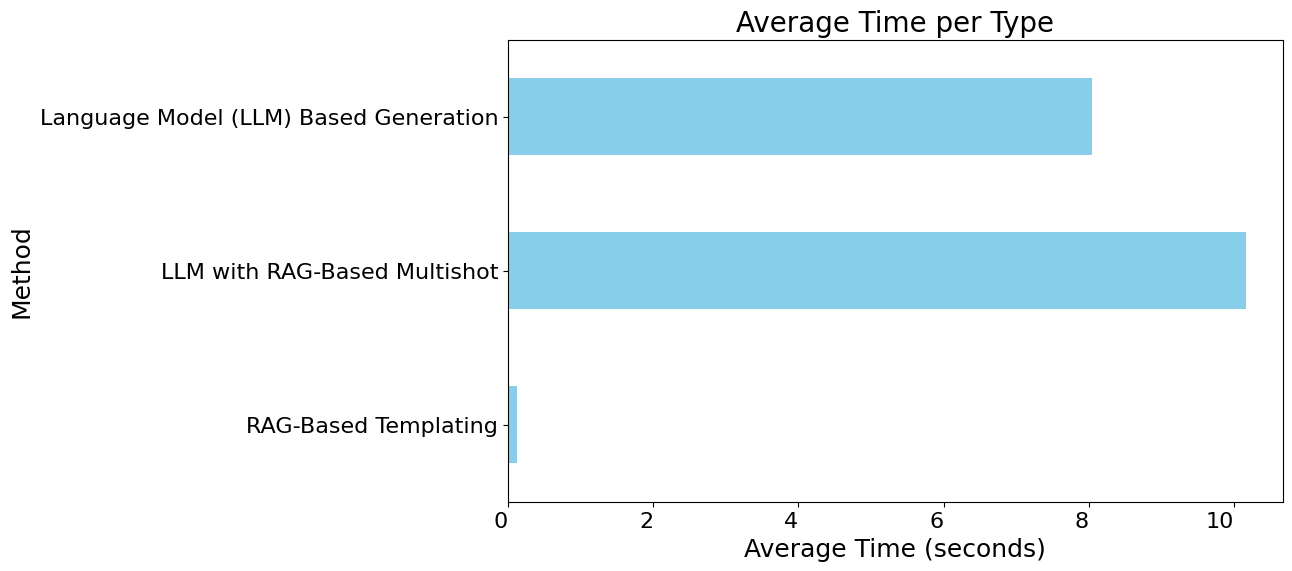}
    \caption{Prompt enhancement times using the three proposed methods.}
    \label{fig:prompt_runtimes}
\end{figure}

\begin{figure}[]
\centering
\includegraphics[width=0.95\textwidth]{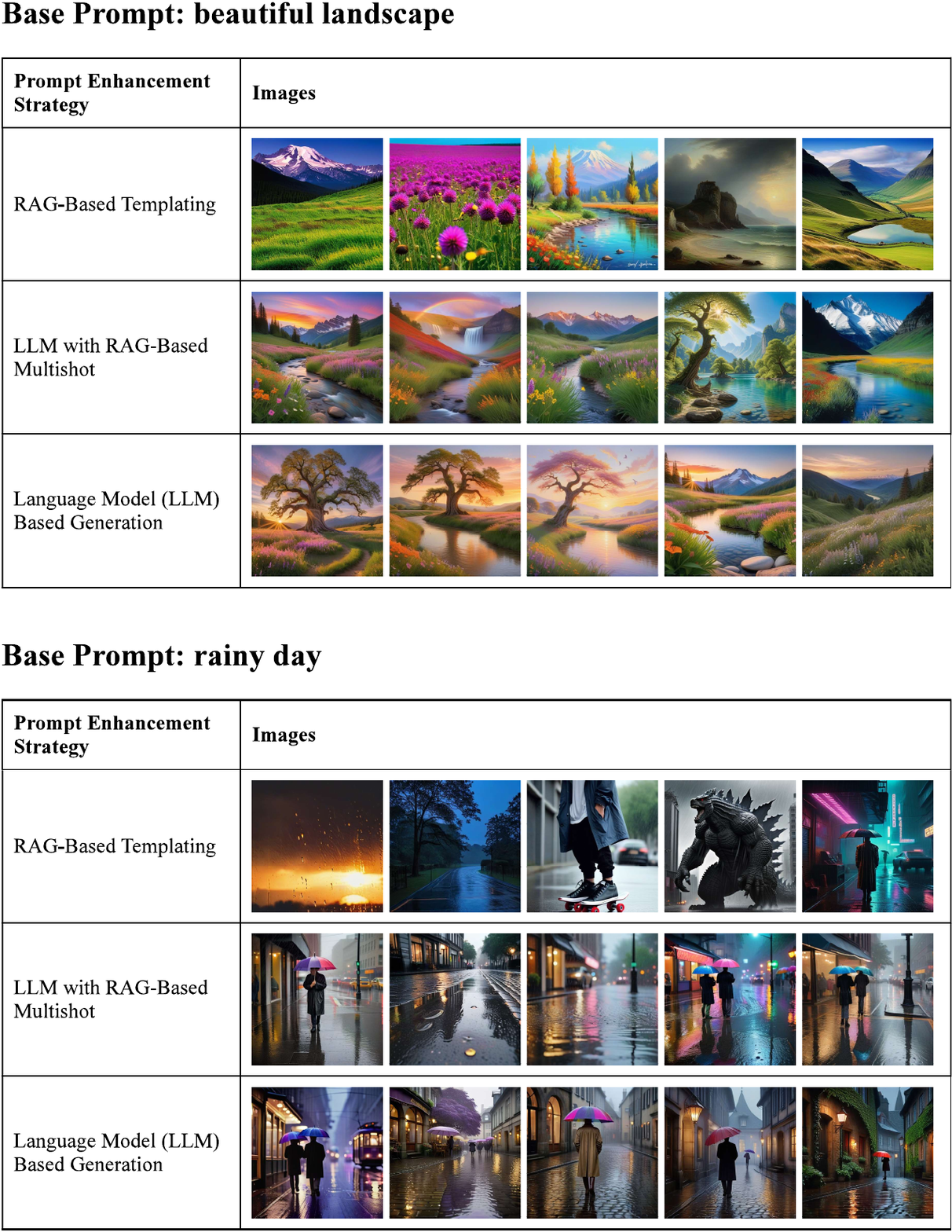}
\caption*{}
\end{figure}
\begin{figure}[]
\centering
\includegraphics[width=0.95\textwidth]{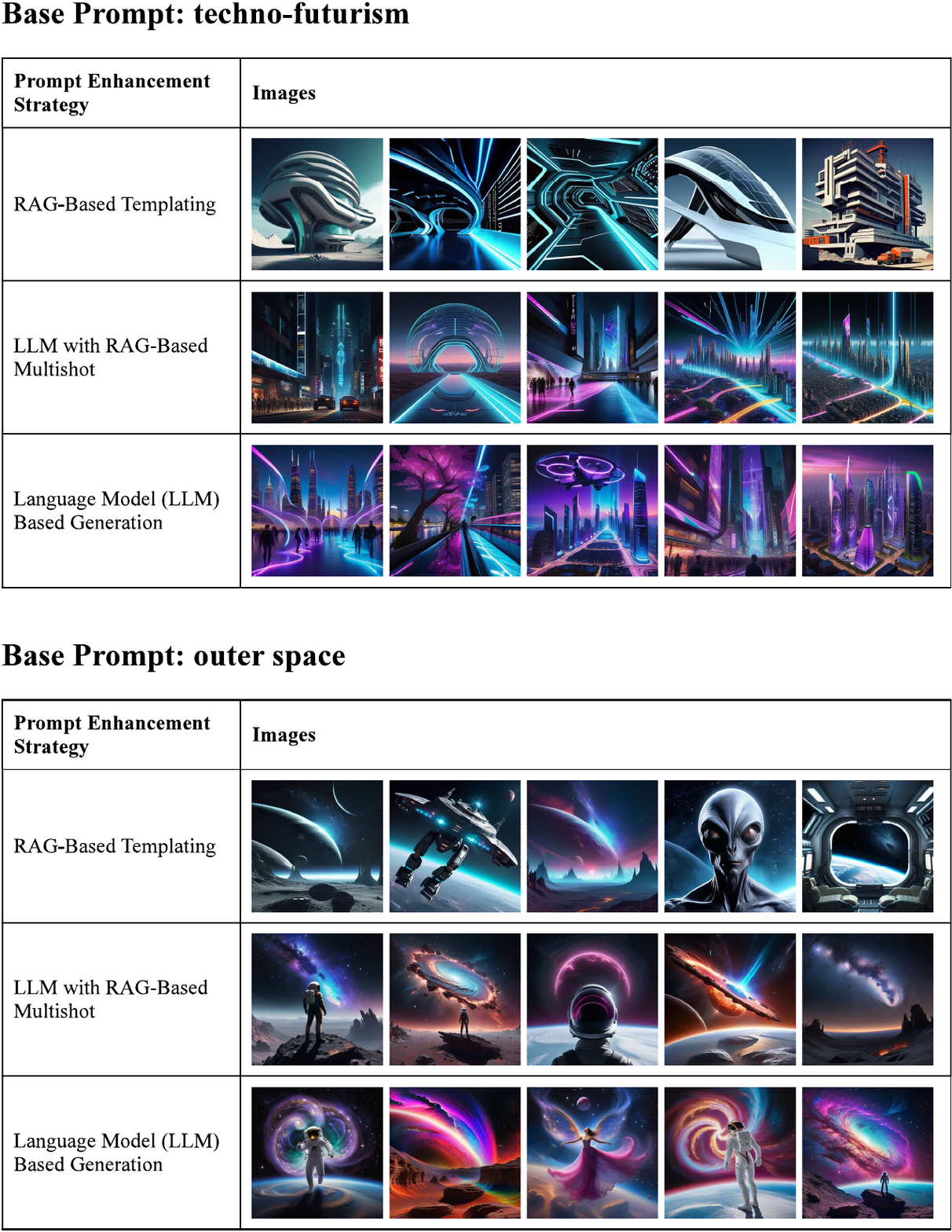}
\caption*{}
\end{figure}
\begin{figure}[]
\centering
\includegraphics[width=0.95\textwidth]{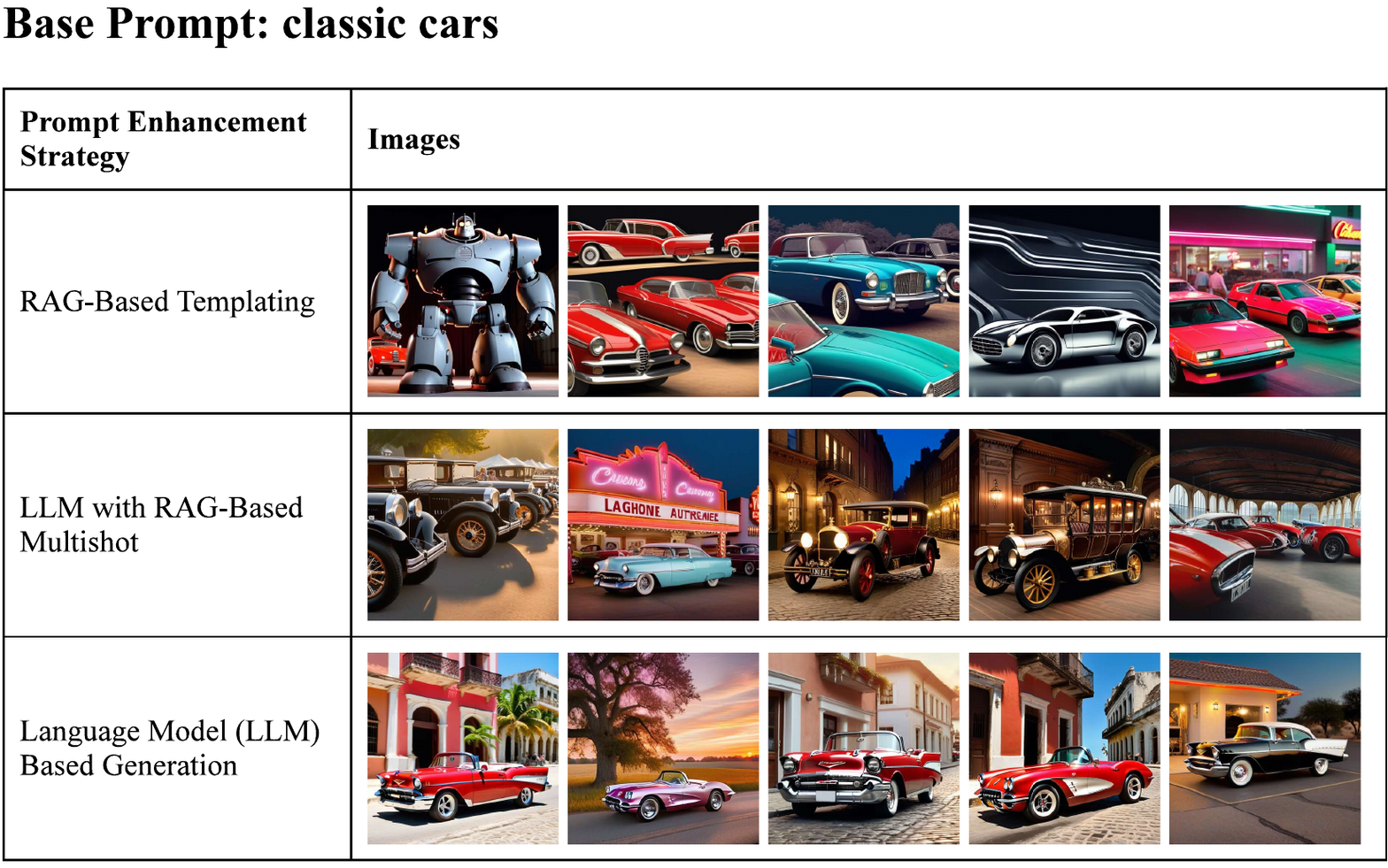}
\caption{Variety of images generated using the three prompt enhancement strategies.}
\label{fig:prompenhance-examples}
\end{figure}

\begin{table}[ht]
\centering
\scriptsize
\noindent\resizebox{\textwidth}{!}{
\scriptsize
\begin{tabularx}{\textwidth}{lXXX}
\toprule
\textbf{Factor} & \textbf{Prompt Templates} & \textbf{LLM Prompt Generation} & \textbf{LLM with RAG Multi-Shot} \\
\midrule
Compute Cost & Low & Medium & High \\
Financial Cost & Low & Medium & High \\
Flexibility & Low & High & High \\
Control Over Prompts & High & High & High \\
Scalability & Low & High & Medium \\
Diversity of Output & High* & High & High \\
Consistency of Output & Low & Medium & Medium \\
Requirement for Pre-existing Data & High (needs a large dataset of templates) & None & High (requires a dataset of prompt examples) \\
Customization Potential & Low to Medium (limited by existing templates) & High (can tailor prompts on the fly) & High (tailored with specific examples in mind) \\
Average Time (sec) & 0.128 & 8.042 & 10.164 \\
\bottomrule
\end{tabularx}
}
\caption{Comparative analysis of prompt enhancement methods}.
\label{tab:dataset_generation_methods}
\end{table}

\cref{fig:prompt_runtimes,fig:prompenhance-examples,tab:dataset_generation_methods} summarizes our experimental results.
Using prompt templates is the cheapest option because the dataset needs to be generated only once, and similarity computation is inexpensive compared to other options. Applying ``rails'' or rules to image generation is straightforward by restricting the base prompts to a limited selection of words and using templates. However, this method can result in odd or mismatched images if there aren't enough templates to cover all the base prompts. For instance, if the base prompt is ``car'' but no templates match ``car,'' the resulting images might be unexpected. This method can lead to greater image diversity due to the combinations of base and enhanced prompts, but if the templates are too few, it can result in low diversity.

Using a Large Language Model (LLM) like GPT-4 for prompt engineering is very expensive. While it generally produces similar results, there's still a chance of obtaining unexpected outcomes. Employing an external model such as OpenAI's GPT necessitates adherence to its content policies, such as restrictions on vulgar prompts. Hosting a model locally is also costly, and smaller models might struggle to generate effective prompts without proper testing.

Incorporating LLM with Retrieval-Augmented Generation (RAG) offers more control over the model, as it draws inspiration from the RAG examples. However, this approach is more expensive because more RAG examples lead to increased token usage. The variety of the output may increase due to the RAG examples providing additional inspiration, but if the RAG examples are too limited, it could also result in less diversity.

\subsection{Upscaling Analysis}

\begin{figure}[]
\centering
\includegraphics[width=0.8\textwidth]{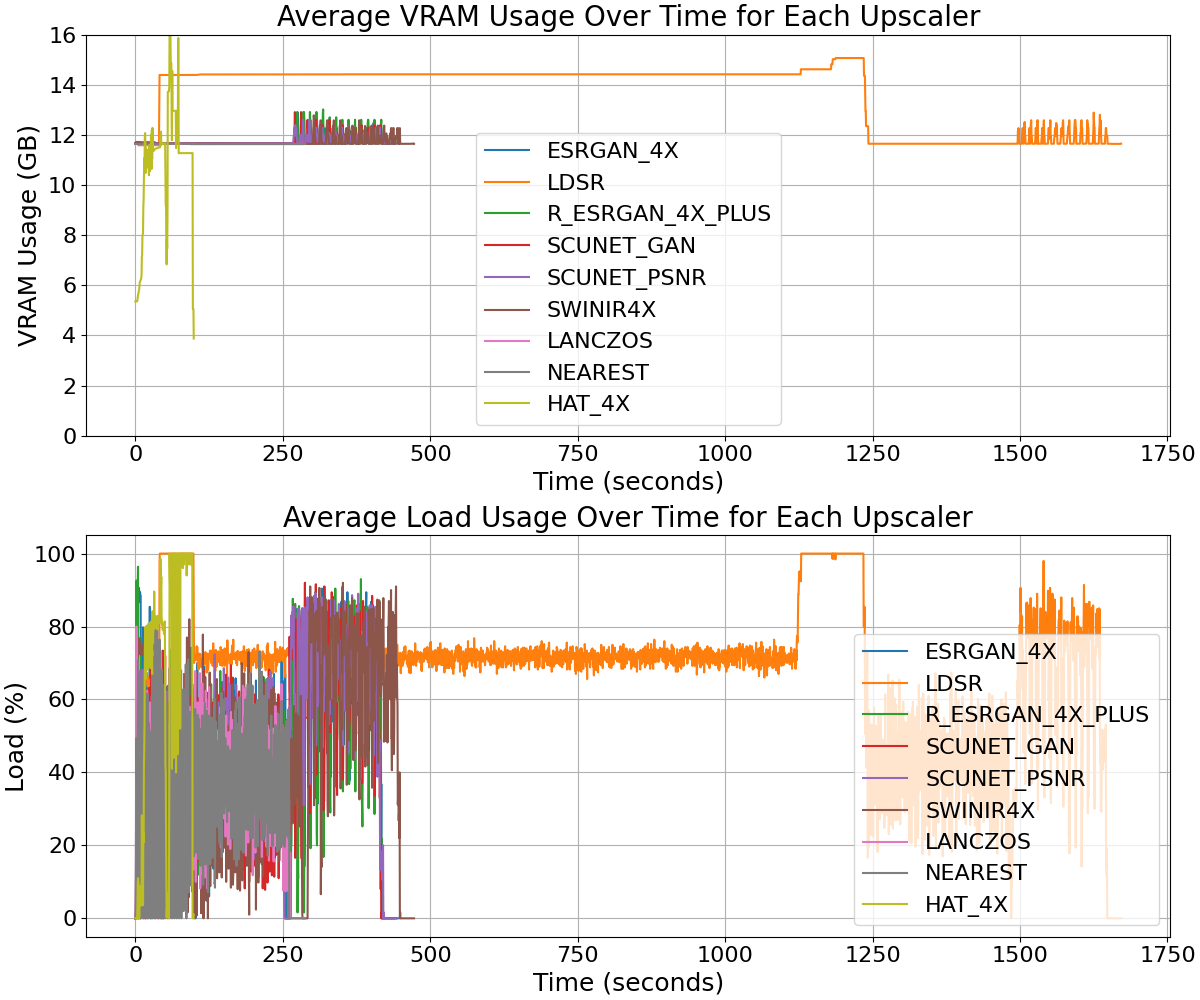}
\caption{GPU usage by each upscaler.}
\label{fig:upscaler-resource}
\end{figure}

\begin{figure}[h!]
  \centering
  
  \begin{subfigure}[b]{0.49\textwidth}
    \centering
    \includegraphics[width=\textwidth]{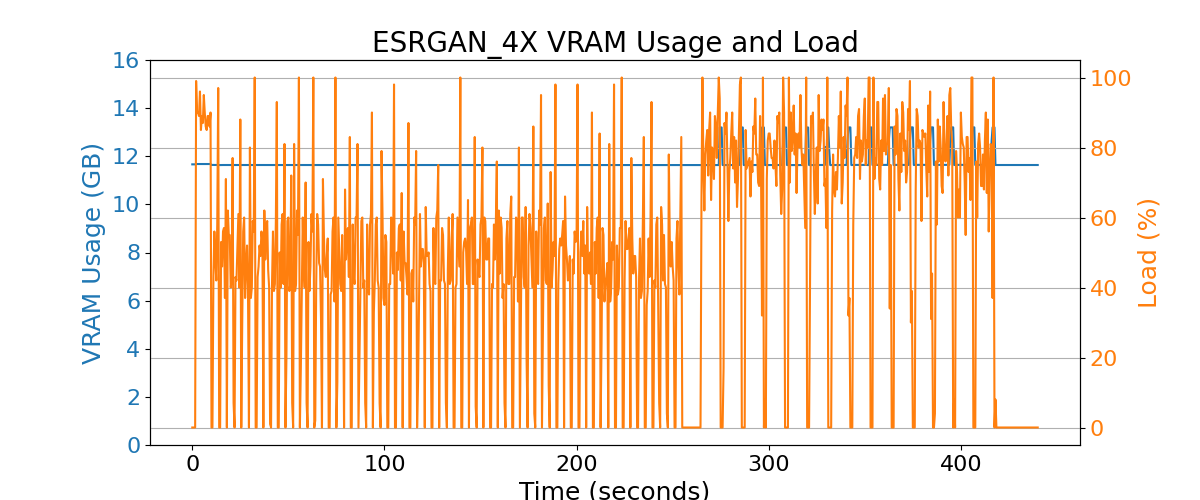}
    \caption{ESRGAN-4X}
    \label{fig:sub1}
  \end{subfigure}
  \begin{subfigure}[b]{0.49\textwidth}
    \centering
    \includegraphics[width=\textwidth]{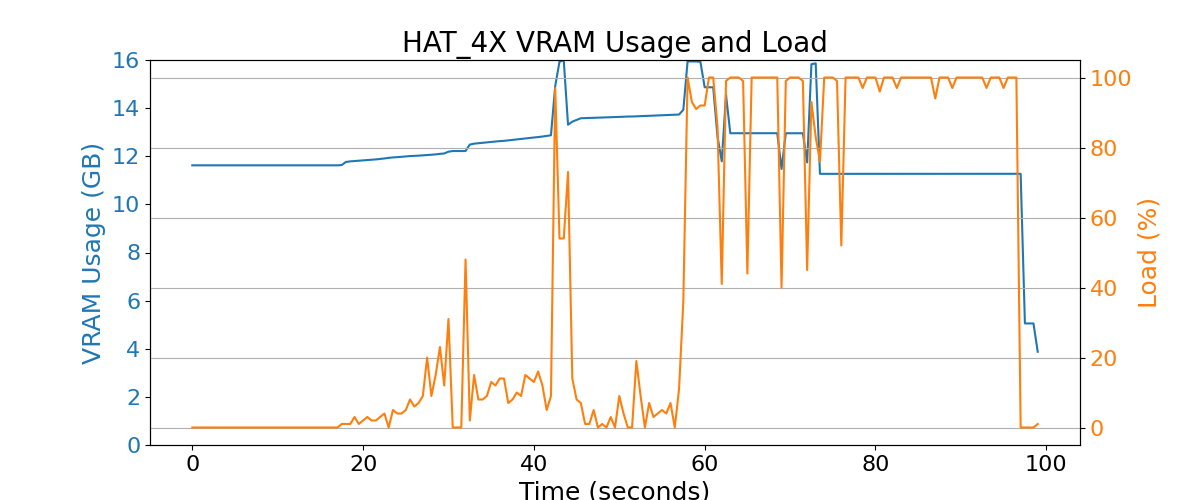}
    \caption{HAT-4X}
    \label{fig:sub2}
  \end{subfigure}

  \begin{subfigure}[b]{0.49\textwidth}
    \centering
    \includegraphics[width=\textwidth]{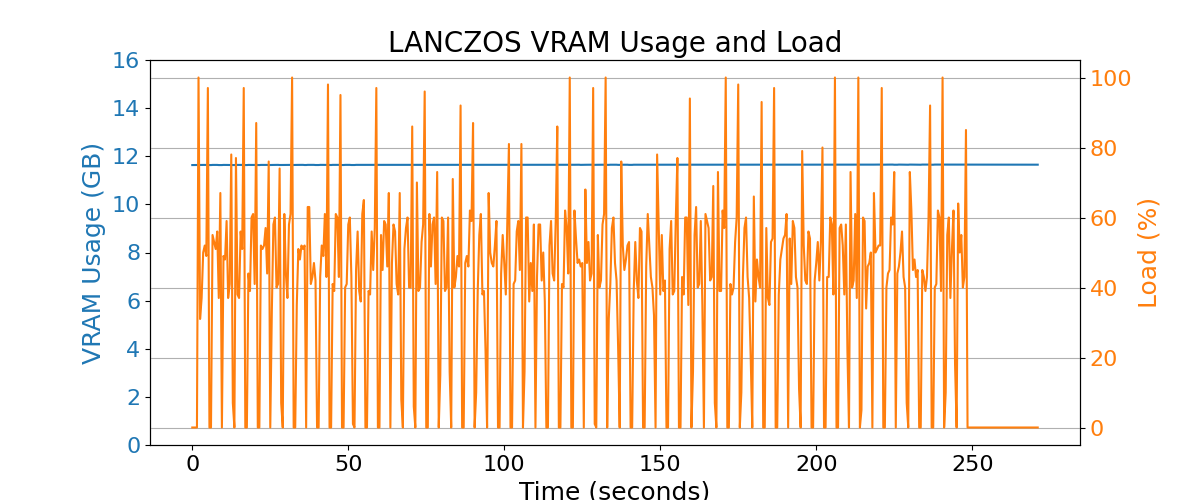}
    \caption{LANCZOS}
    \label{fig:sub3}
  \end{subfigure}
  \begin{subfigure}[b]{0.49\textwidth}
    \centering
    \includegraphics[width=\textwidth]{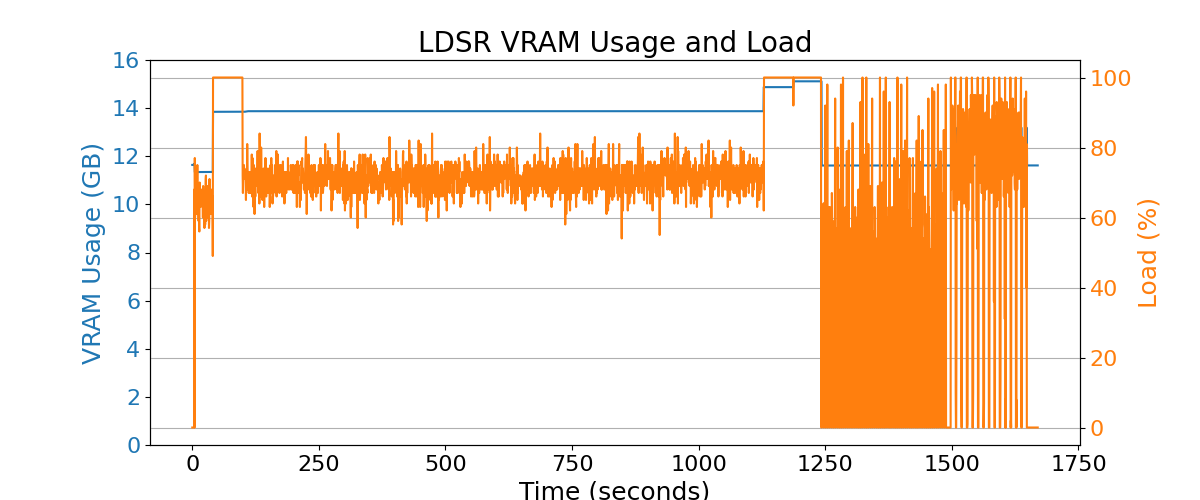}
    \caption{LDSR}
    \label{fig:sub4}
  \end{subfigure}

  \begin{subfigure}[b]{0.49\textwidth}
    \centering
    \includegraphics[width=\textwidth]{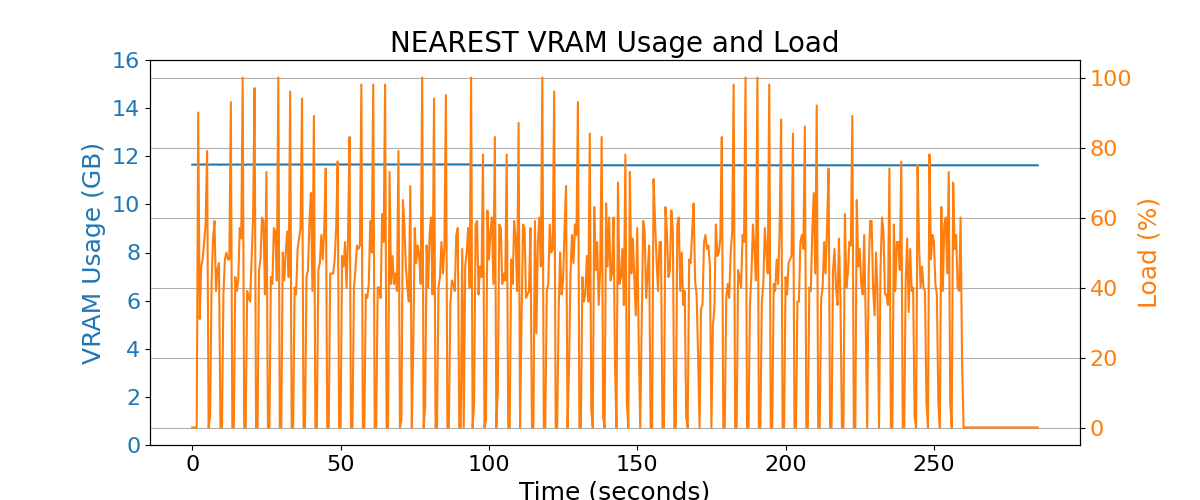}
    \caption{NEAREST}
    \label{fig:sub5}
  \end{subfigure}
  \begin{subfigure}[b]{0.49\textwidth}
    \centering
    \includegraphics[width=\textwidth]{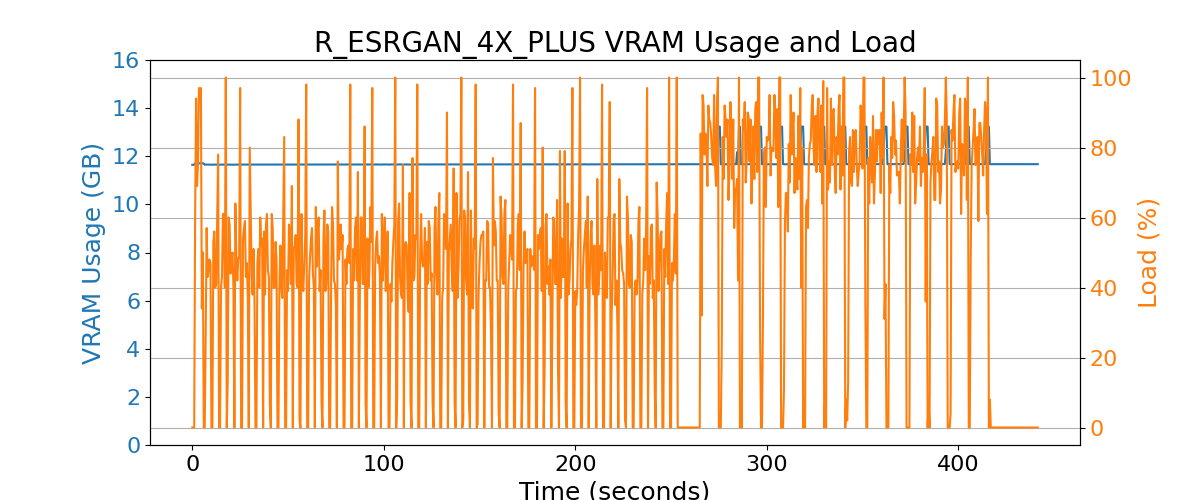}
    \caption{R-ESRGAN-4X-PLUS}
    \label{fig:sub6}
  \end{subfigure}
  
  \begin{subfigure}[b]{0.49\textwidth}
    \centering
    \includegraphics[width=\textwidth]{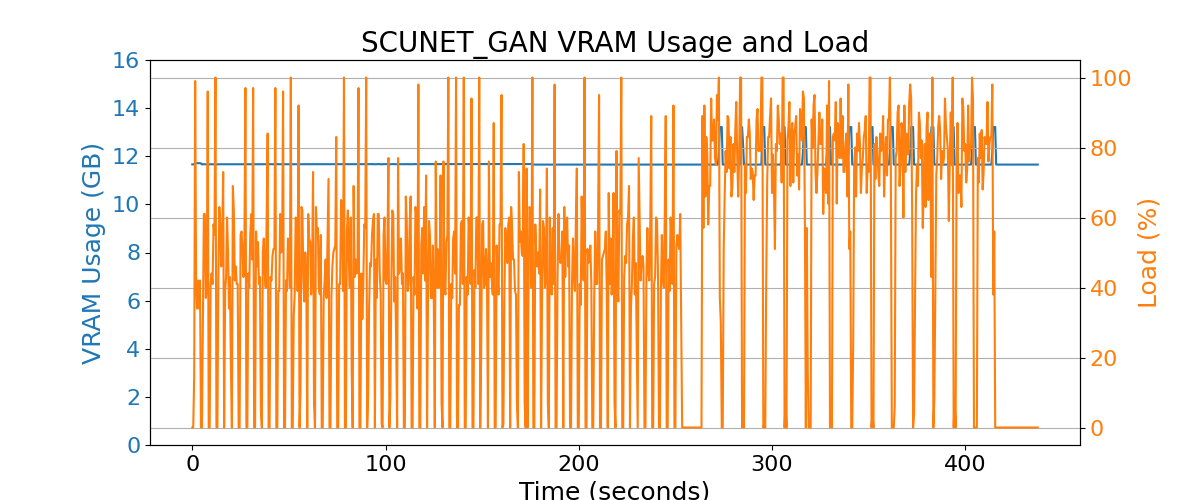}
    \caption{SCUNET-GAN}
    \label{fig:sub7}
  \end{subfigure}
  \begin{subfigure}[b]{0.49\textwidth}
    \centering
    \includegraphics[width=\textwidth]{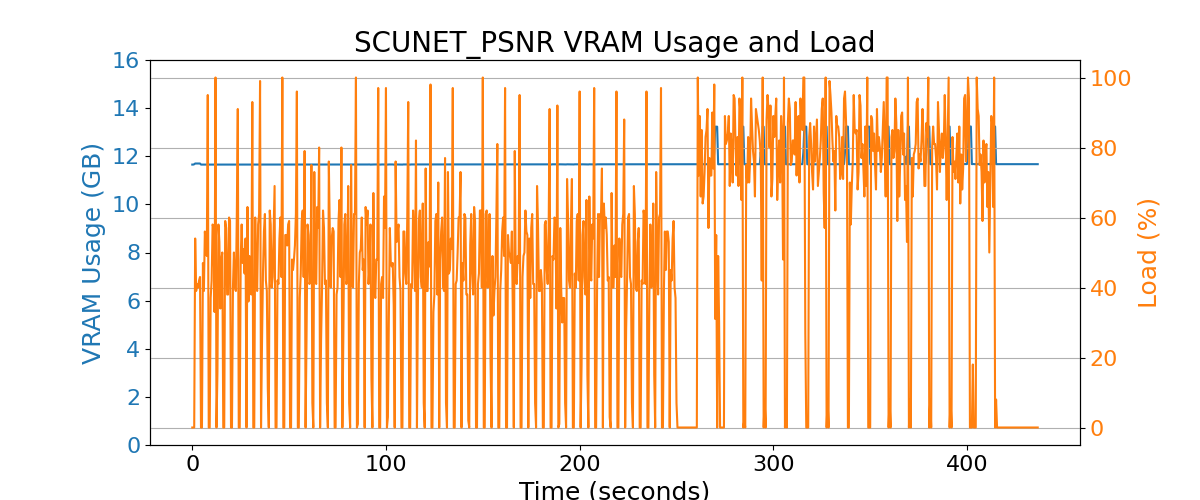}
    \caption{SCUNET-PSNR}
    \label{fig:sub8}
  \end{subfigure}

  \begin{subfigure}[b]{0.49\textwidth}
    \centering
    \includegraphics[width=\textwidth]{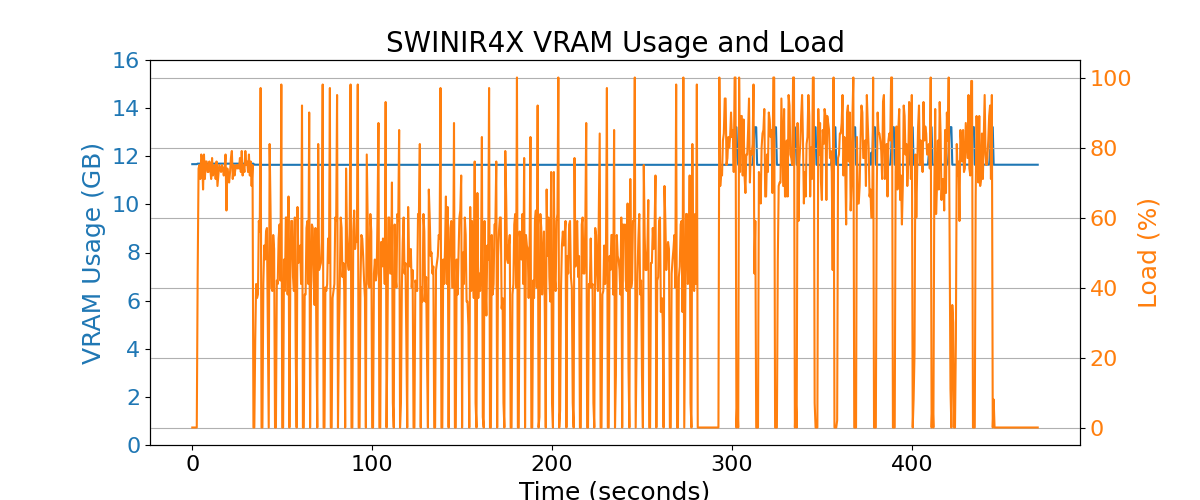}
    \caption{SWINIR4X}
    \label{fig:sub9}
  \end{subfigure}
  
  \caption{Blue lines represent VRAM usage, and orange lines indicate GPU load for various upscaling models.}
  \label{fig:series_subfigures}
\end{figure}

\begin{figure}[H]
\centering
\includegraphics[width=0.8\textwidth]{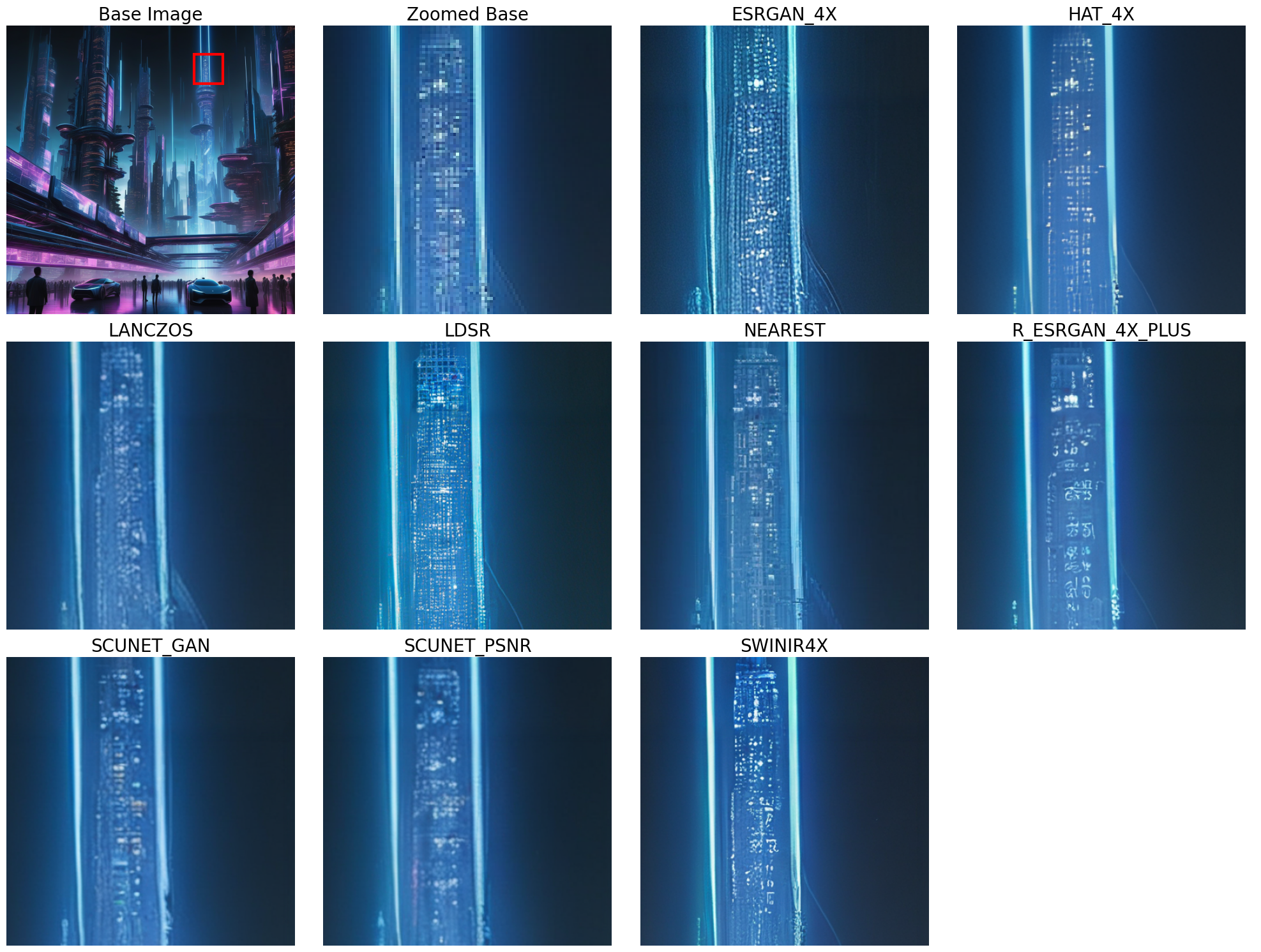}
\caption{Examples of upscaled image quality.}
\label{fig:upscaled-examples}
\end{figure}

\begin{figure}[htbp]
\centering
\begin{subfigure}{.32\textwidth}
  \centering
  \includegraphics[width=\linewidth]{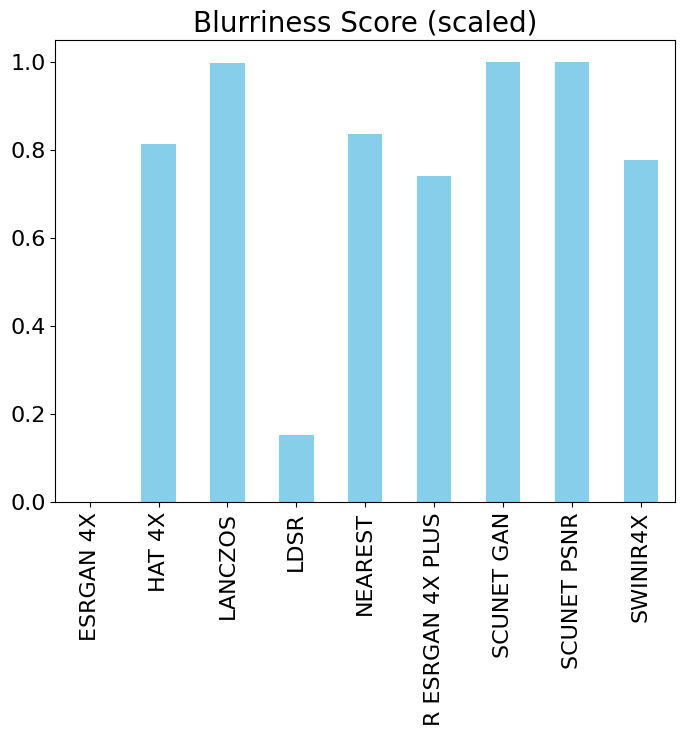}
  \caption{Blurriness}
  \label{fig:blurriness}
\end{subfigure}%
\hfill
\begin{subfigure}{.32\textwidth}
  \centering
  \includegraphics[width=\linewidth]{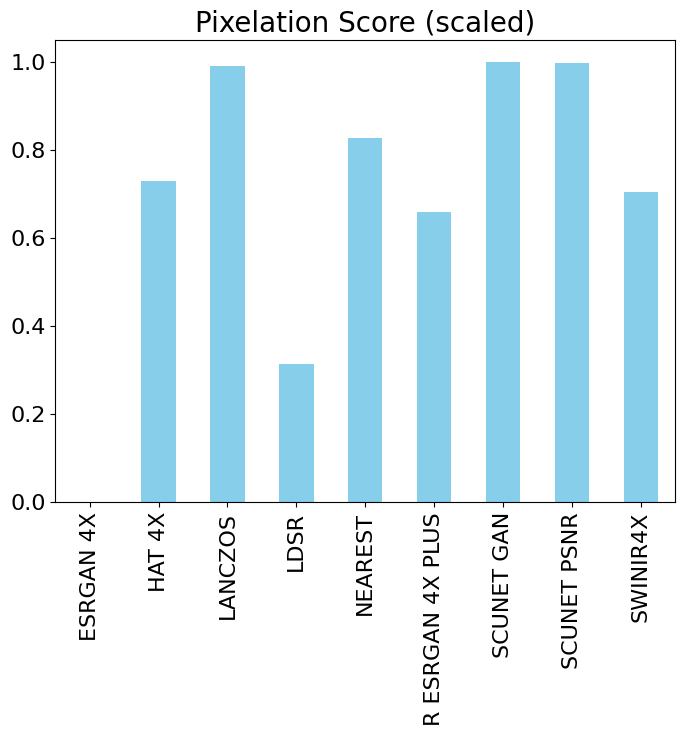}
  \caption{Pixelation}
  \label{fig:pixelation}
\end{subfigure}%
\hfill
\begin{subfigure}{.32\textwidth}
  \centering
  \includegraphics[width=\linewidth]{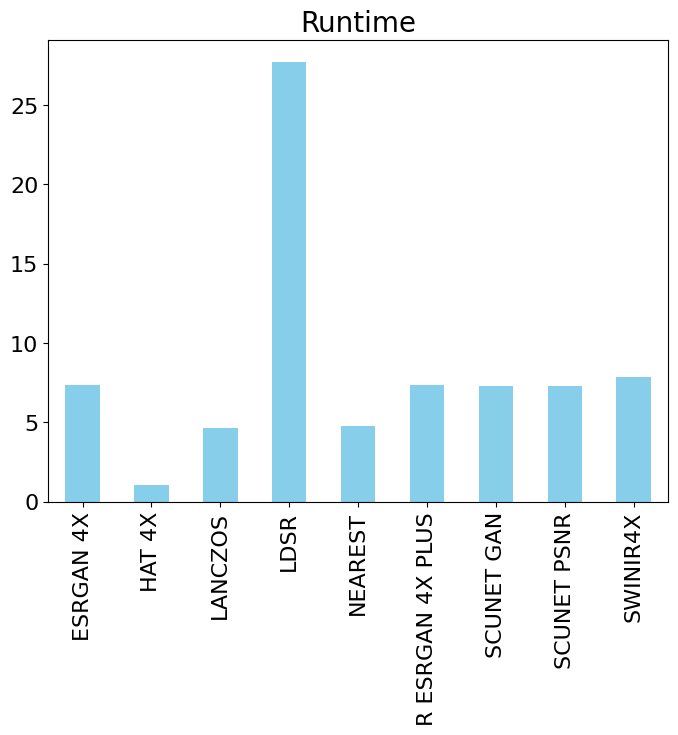}
  \caption{Runtime}
  \label{fig:time_data}
\end{subfigure}
\caption{Different quality factors measured on upscaled images.}
\label{fig:upscaled-quality}
\end{figure}

We analyzed several examples of upscaled AI-generated artwork images (\cref{fig:upscaled-examples,fig:upscaled-quality,table:upscaler_run_time}). In the competitive landscape of image upscaling, the efficiency of the upscaling process is critical for commercial applications. The HAT-4x model emerges as a particularly compelling option due to its superior speed, outperforming other models in time to upscale while maintaining satisfactory quality. This balance between speed and quality makes it a frontrunner for scenarios where turnaround time is a decisive factor. Meanwhile, ESRGAN 4X stands out for its exceptional quality, achieving the highest pixelation score indicative of detailed image enhancement, along with the lowest blurriness score, ensuring images remain sharp and clear post-upscaling.

On the other hand, models with higher blurriness scores are generally inclined to smooth images, which can sometimes be at the expense of detail and sharpness. This is not the case with ESRGAN-4X, which excels at clarity, or with ScuNET, which tends to produce very sharp, albeit sometimes overly pixelated, images. LDSR, despite its longer run time, fails to lead in either pixelation or blurriness, placing it lower in the hierarchy of preferred models. Additionally, models such as R-ESRGAN-4x PLUS and SwinIR-4X have a notable propensity to misinterpret image details as text, a characteristic reflected in their closely matched pixelation and blurriness scores, and an aspect that may be a drawback for certain applications.

\clearpage
\begin{figure}[H]
\centering
\includegraphics[width=0.8\textwidth]{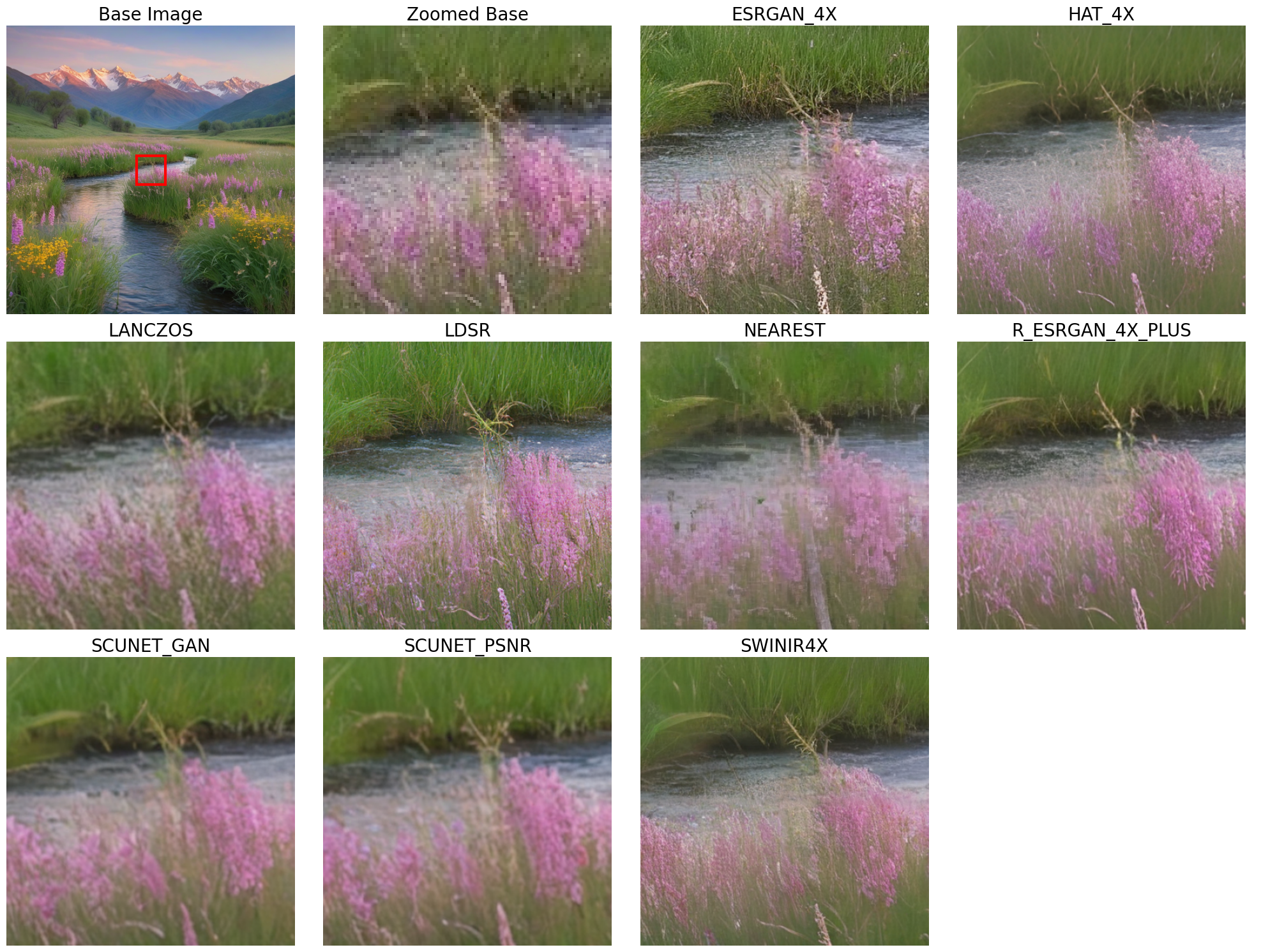}
\caption*{}
\end{figure}
\begin{figure}[H]
\centering
\includegraphics[width=0.8\textwidth]{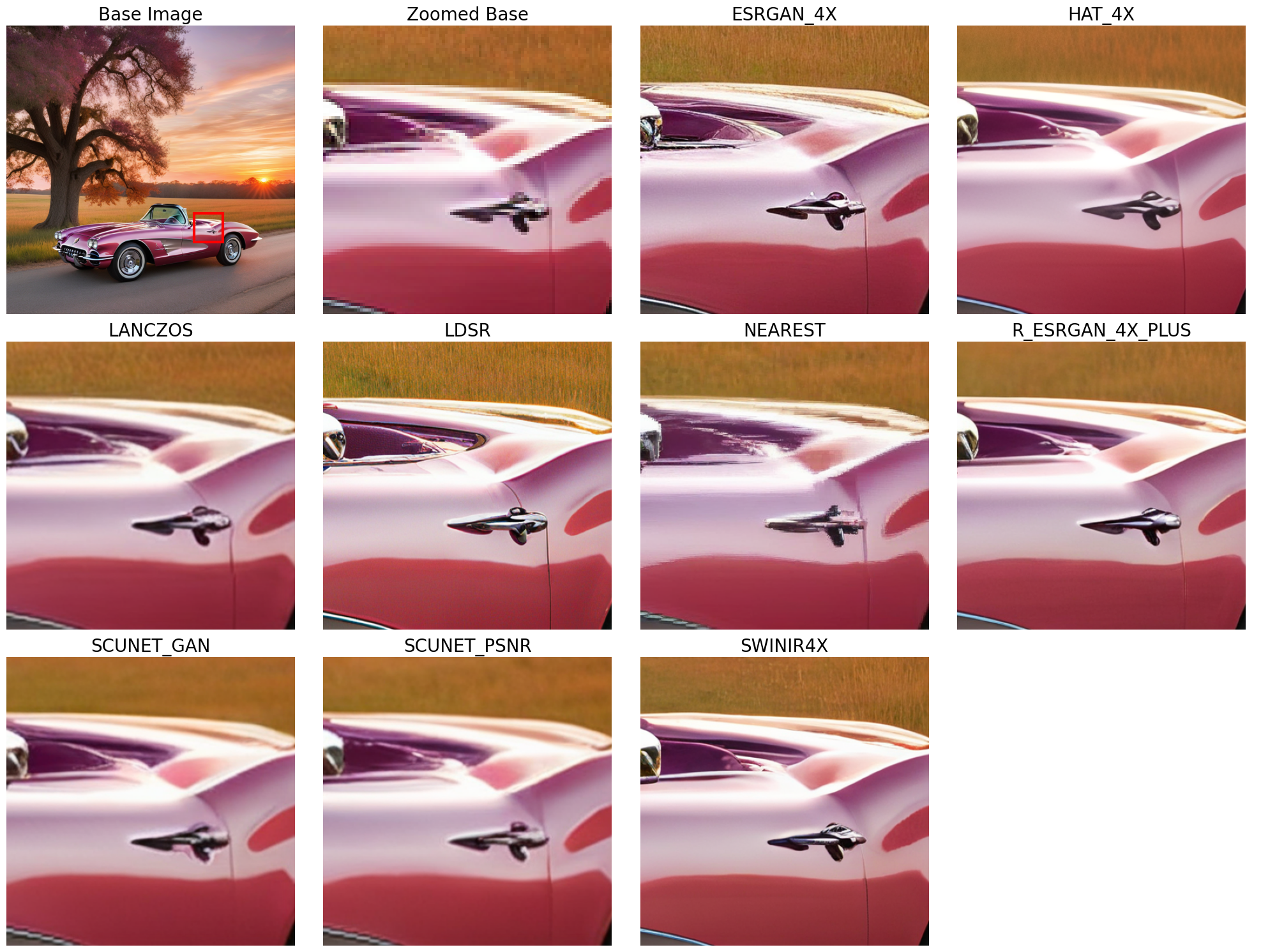}
\caption*{}
\end{figure}
\begin{figure}[H]
\centering
\includegraphics[width=0.8\textwidth]{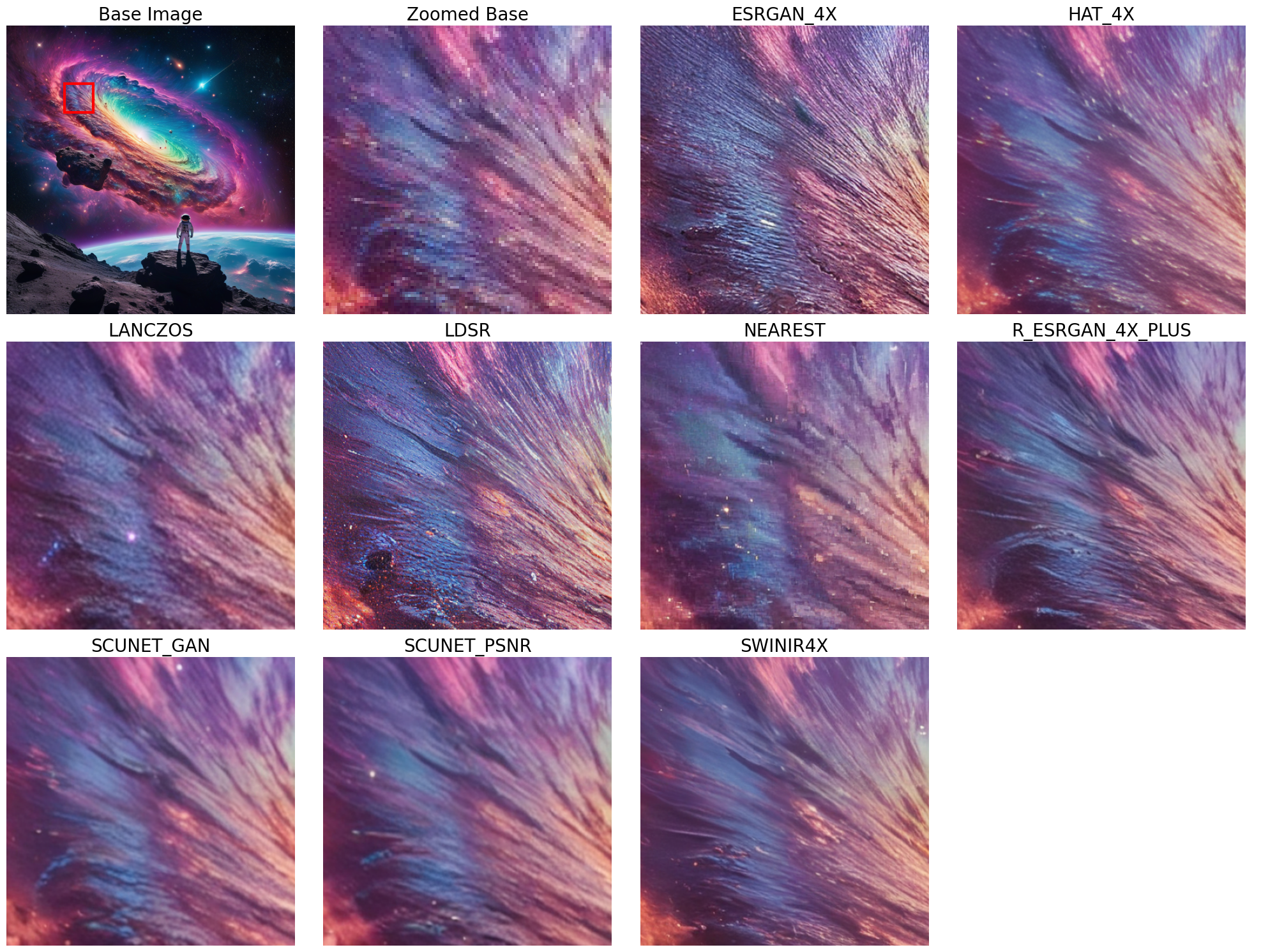}
\caption*{}
\end{figure}
\begin{figure}[H]
\centering
\includegraphics[width=0.8\textwidth]{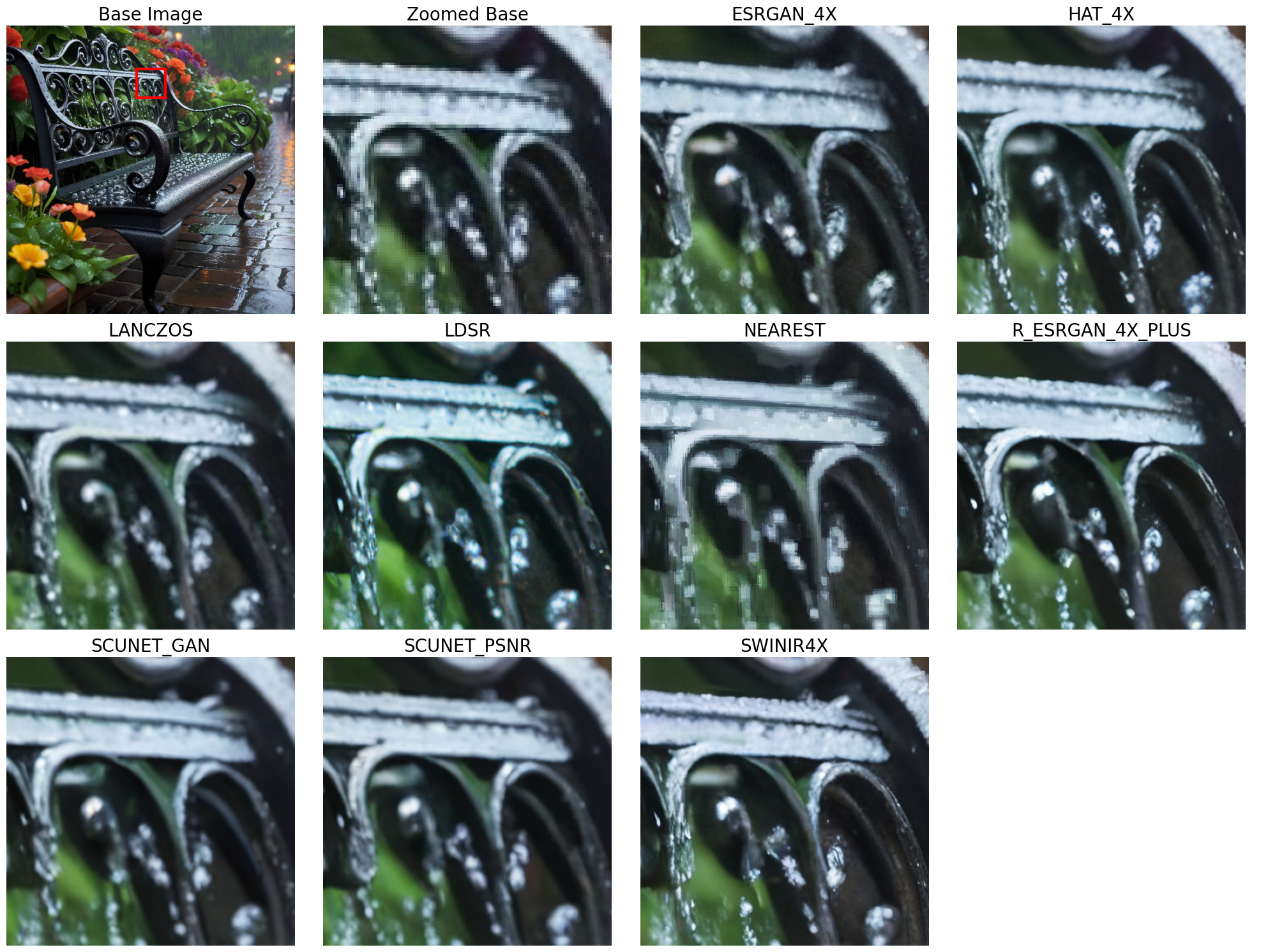}
\caption*{}
\end{figure}

\begin{table}[H]
\centering
\begin{tabular}{cc}
\hline
\textbf{Upscaler} & \textbf{Average Run Time (minutes)} \\ \hline
ESRGAN\_4X & 7.325 \\
LDSR & 27.743 \\
R\_ESRGAN\_4X\_PLUS & 7.36 \\
SCUNET\_GAN & 7.273 \\
SCUNET\_PSNR & 7.277 \\
SWINIR4X & 7.85 \\
LANCZOS & 4.65 \\
NEAREST & 4.763 \\
HAT\_4X & 1.052 \\ \hline
\end{tabular}
\caption{Average run time of each upscaler.}
\label{table:upscaler_run_time}
\end{table}

\section{Conclusion}

We present a new way to make high-quality art using AI, specifically for things like posters and printed products that need to be really clear and detailed. The main focus is on solving two big problems: making the process of telling the AI what to create (prompt engineering) easier and improving the quality of the images so they are sharp enough for large prints. We mixed techniques for crafting these instructions and for making the images look better when they are made bigger. Our experiments show that this approach works well, making it easier for everyone, from everyday people to professional designers, to create cool AI art that looks good even on big canvases. This study is a step forward in making AI art more usable and enjoyable for a variety of purposes, like decorating spaces or making unique products.

\bibliographystyle{unsrt}
\bibliography{sample}

\end{document}